# Understanding Design Fundamentals:
# How Synthesis and Analysis Drive Creativity, Resulting in Emergence[1]

V.V. Kryssanov[†], H. Tamaki, and S. Kitamura

**Abstract**

This paper presents results of an ongoing interdisciplinary study to develop a computational theory of creativity for engineering design. Human design activities are surveyed, and popular computer-aided design methodologies are examined. It is argued that semiotics has the potential to merge and unite various design approaches into one fundamental theory that is naturally interpretable and so comprehensible in terms of computer use. Reviewing related work in philosophy, psychology, and cognitive science provides a general and encompassing vision of the creativity phenomenon. Basic notions of algebraic semiotics are given and explained in terms of design. This is to define a model of the design creative process, which is seen as a process of semiosis, where concepts and their attributes represented as signs organized into systems are evolved, blended, and analyzed, resulting in the development of new concepts. The model allows us to formally describe and investigate essential properties of the design process, namely its dynamics and non-determinism inherent in creative thinking. A stable pattern of creative thought – analogical and metaphorical reasoning – is specified to demonstrate the expressive power of the modeling approach; illustrative examples are given. The developed theory is applied to clarify the nature of emergence in design: it is shown that while emergent properties of a product may influence its creative value, emergence can simply be seen as a by-product of the creative process. Concluding remarks summarize the research, point to some unresolved issues, and outline directions for future work.

<u>Keywords</u>: Engineering design, creativity, semiotics, emergence.

† The corresponding author. Inquiries should be sent to Dr. Kryssanov V.V., Center for Information and Multimedia Studies, Kyoto University, Kyoto 606-8501, JAPAN. Fax: +81-075-753-9056, e-mail: kryssanov@mm.media.kyoto-u.ac.jp

---

# 1    Background

## 1.1    Engineering design

The word 'design' has at least three important meanings: as a process, an object, and as a discipline. Engineering design is associated with human activities aimed at and ultimately resulting in the development of an artifact – the product. These activities take place throughout the product life cycle and typically include (but are not limited to) the following:

- the identification of consumers' needs;
- the conceptualization of design requirements to meet the needs;
- the transformation of the requirements into performance and function specifications;
- the elaboration of cost, resource, and other critical constraints;
- the mapping and converting of the specifications into feasible design solutions;
- the optimization of the solutions;
- the documenting of the developed concept and the prototyping of the product;
- the manufacture of the product;
- the implementation of various control, logistics, and marketing procedures;
- providing for maintenance and other after-sale services of the product; and
- obtaining and utilizing feedback concerning the product's utility, operation, and disposal, as well as its market value.

The design process generates a design – an explicit representation of an engineering system in a design (abstract) space. A design space is created to describe engineering system concepts and concept attributes in terms of a design language. Each concept represents a class of engineering entities, where a particular entity can be selected by assigning specific attributes to the concept. Because a design is, first of all, an information object, the design process necessarily involves and is based on various information processing activities.

Besides, the design process is a socio-economic and, by its very nature, mental phenomenon. One possible definition of design as a scientific discipline would be a study of the thought process comprising the creation of an artifact in a given (social, technical, economical, etc.) environment.

The design process is frequently thought of in terms of several sequential phases: conceptual design, preliminary or embodiment design (layout), and detailed design. Design as an information object cannot be created 'from nothing'. Gathering and analyzing marketing information, information about the production company, as well as relevant knowledge and experience usually precede the design first phase and provide an information bulk necessary to start the development of a new product concept. During conceptual design, the product desired functionality is determined, potential design solutions and the corresponding performances are evolved, the expected expenses (product costs) are estimated, restrictions on the potential solutions are found, and a product general specification consisting of a description of the design concepts along with function and behavior constraints is defined. The specification gives rise to an initial design space. The next design phase stipulates realizing the product components' structure. During embodiment design, the product specification is refined by eliminating design alternatives infeasible in general or impracticable under given available recourses, and searching across the design space for local optima (in terms of concepts) with least risks and best performances is accomplished. There may be defined more than one optimization criterion for the product so that some of the criteria may not always be satisfied. This may require re-defining the product specification, back to conceptual design. Once the design process has come to its detailed phase, the design space is elaborated, specialized, and explored in pursuit of discovering a single high-performance global solution for the product functional requirements and structure. Ensuring the design efficiency may necessitate several iterations to be done over the preliminary and detailed design phases until an overall design solution is found. Another cause for recurring re-designing of the product can be the need to respond to shortcomings of the design detected in the later stages of the product life cycle. Checking the design integrity and consistency as well as conformance to the current specification closes each of the loops in the design process. Note that in most cases, designing involves multiple individuals over the different phases and at different times that introduces a strong collaborative aspect into the above discussion.

The gist of conceptual design as a cognitive activity is realizing the objectives, and functionality and properties satisfying the objectives for a new product concept. It would take a long time and a great deal of effort to systematize, search through, and analyze a large amount of multi-disciplinary information to create a specification of a new product. In practice however, the designer's work at this phase is often based on proficiency, inspiration, and creativity rather than on routine search and analysis. This is due to the human ability to draw and utilize metaphors and analogues from past experience and elicit and apply knowledge not limited to one domain, thus synthesizing hypotheses about design possible solutions that allow for localizing and specializing the originally immense (in terms of possible states) design space. Creativity and human synthetic abilities play the prime rôle during conceptual design.

In embodiment design, the focus of human design activities gradually shifts from synthesis of the product concept to exploration of the design potential solutions. A set of feasible configurations of the product is defined through examination of various combinations of product components, their constraints and interactions, and the available resources and technologies to ensure that all of the components can principally be manufactured and will work in the context of the product as a whole. For each configuration, most promising (e.g. in terms of the efficiency) design directions for every product component are determined, where a direction corresponds to a certain region in the multi-dimensional design space. Preliminary design may be understood as multi-criteria optimizing search across a set of disjoint regions of the design space. A majority of decisions made during embodiment design are based upon the domain empirical knowledge and concern with adaptation and use of design solutions obtained in the past for similar problems. The rôle of creativity becomes less conspicuous, but innovation and proficiency are urgently needed at this phase of the design process.

Detailed design is to develop to the highest possible efficiency all of the product's component elements, to flesh out the necessary drawings, technical details, specifications, and tolerances that will allow the product to be manufactured. No major changes should occur in the design at this phase, and the degree of uncertainty is far less than ever before. However, there still remain numerous design alternatives to be brought to a single solution that should meet several optimization criteria, such as minimal cost, maximal lifetime, minimal disposal effort and environmental impact, and the like. Cognitive processes behind detailed design are, in the main, of search, where there is little room for innovation and almost no room for synthesis and creativity.

**1.2     Computer-aided design**

Over the last two decades, several computer-based design methodologies have been proposed to increase the effectiveness and achieve better control over different phases of the design and production processes. Evbuomwan *et al*. [1] surveyed various design philosophies, methodologies, and methods. Willemse [2] and Monteiro *et al*. [3] made an extensive analysis of computer-based design support models and systems. Recent works by Yazdani and Holmes [4] and Váncza [5] have shown that since the early days of design computerization, no principal breakthrough has come about in applying computer-aided design tools and understanding their rôle in the design process.

Traditionally, a computer-based design methodology covers technical and organizational aspects of the design process, providing with systems, methods, and procedures to support design routine activities, such as documentation, storage, and translation of the design results. The attention of contemporary design computer applications is on the later phases of the design process, while the early phases are still poorly automated and receive little information support [6]. There are reasons to think that the absence of a design theory, which would coherently explicate the whole design process in a scientific and unambiguous way, is the main predicament for the development of more sophisticated computer tools capable of assisting human designers in their non-routine activities [7].

In the past, there have been few attempts to introduce theoretical principles for design, which would be suitable for computer treatment, derived from: generalization of factual data, empirical knowledge, and personal experience (e.g. [8, 9]), application of some philosophical and logical hypotheses to best engineering practice (e.g. [10, 11]), or adaptation of ideas from other disciplines, such as control theory, functional analysis, Shannon-Weaver information theory, etc. (e.g. [12]). Besides, a number of computer-aided design systems, which reflect just the developer's (usually situated and subjective) understanding of the design activities, have been brought to the market and installed at production companies, interfering in and, sometimes unpredictably, affecting the process of designing. Based on an analysis of the literature, Figure 1 gives an integral framework representing the horizon of modern computer-aided design methodologies.

The framework incorporates features of descriptive (by categorizing design activities) and prescriptive (by ordering these activities) models of the entire design process, excluding the technical stages of documenting of design rationale, methods, and results. There are three 'information spaces' evolving in the design process: the objectives (requirements) space, the solutions (alternatives) space, and the models space. The process is seen as a progressive but not necessarily monotonic transition from a set of available inputs (the requirements) to a set of desired outputs (the alternatives) through analysis and exploration. The transition is coordinated with the design activities over the design phases. The solid arrows show the mappings between the spaces, and the dashed arrows indicate the major actions (the outward) and their expected results (the inward).

It is understood from the figure that although there is no strictly ordered progression from one design activity to another (rather, there are a few cycles, including the 'design main cycle', associated with the selection of design solutions under modeling, exploration, and optimization), the model postulates the linear character of the design information flow. That is, once the requirements have been formulated and fixed, they are mapped, refined, step-by-step interpreted, analyzed, modified, and detailed in a variety of domains, such as simulation domain, manufacturing domain, consumer domain, etc. This pattern is valid for traditional sequential design approaches but also in the context of the recently popular design paradigms, such as concurrent engineering, design for X, parallel and condensed design development, and the like [4].

Contemporary computer-aided design methodologies deal with a model as the main information-theoretic object, and their rôle is usually seen as to define mappings between design models represented in different information spaces (e.g. by specifying protocols for the transfer of the information) and to guide exploration of these models (e.g. by prescribing

-6-

appropriate analysis and optimization procedures).

However, it is now well recognized that in the beginning of the design process, there exists no model but only a partial and intuitively understood image of the product that tends to change drastically and dynamically at the later phases [6, 13]. Even a completed design is the description of not what is the product, but what it might be. Furthermore, there is non-determinism intrinsic to human designing that, perhaps, cannot be addressed by any of the design approaches based on the use of closed (i.e. 'from top to bottom' completed) models, whether of information or of knowledge. All this together with the poor understanding of fundamentals of the design creative process prevents from applying computers and information technologies effectively during early, conceptual design.

## 1.3    Motivations and key assumptions

Computer-aided support in design usually builds on results of an analysis of the design activities from a functional, information processing, economic or, most recently, ethnomethodological perspective, and it is aimed at increasing the design efficiency (in a broad sense). Automating organizational, technical, and analytical aspects presently dominates over any other possible applications of computers in the design process. One of the designer's most important functions is, however, to synthesize, to create new things. The creative process is (at least partially) unconscious; thought is impossible to observe and difficult to explain. Even if it is sometimes claimed that a particular tool is able to foster and enhance the designer's creativity, it is still unclear what is meant by 'enhancing creativity,' and whether computers can deal with creativity at all. The latter is a fundamental question subject to debate [14, 15]. Any computational theory of design has, nevertheless, to address design synthesis and creativity in a pragmatic manner. One possible way for building such a design theory would be through understanding and modeling the elementary cognitive mechanisms underlying creative thought.

The focus of the presented research is on the development of a computational model of the creative process, while our ultimate goal is to develop a scientific theory of design. It is also a purpose of our study to identify key characteristics of next generation methodologies of computer-aided design.

This work draws upon the ideas of semiotics – a science about signs, signification, and sign systems. Semiotics provides a powerful and elegant tool for the investigation of thought and evolutionary processes involving products and their environments of operation [16]. Semiotics deals with an '…action, or influence, which is, or involves, an operation of three subjects, such as a sign, its object and its interpretant, this tri-relative influence not being in any way resolvable into an action between pairs' [17], and it studies how signs mediate meaning through the processes of semiosis, in which something functions as signs. Semiosis is classically (in Peircean terms [17]) defined on and affects the *signifier* (the sign vehicle or token; usually a sign), the *signified* (the designatum, e.g. an object, whether physical or abstract), and the *interpretant*, i.e. the sense made of the sign or that, which follows semantically from the process of interpretation (also, see [18]).

Creative thinking, just like any other cognitive process, is based on the representation and processing of information (presumably encoded in signs) in the mind [19]. Hence, the design creative process can be associated with and studied as a process of semiosis, where concepts and ideas represented as signs organized into systems are continually blended, analyzed, and developed [20]. Furthermore, as any theory, language, or model naturally forms a specific sign system, finding an appropriate computational interpretation of basic semiotic notions would allow for establishing a common computational ground for different design theories.

## 2       Human creativity: a survey

Creativity is a complex phenomenon, and its study usually includes four aspects: the *creative process*, the *creative agent*, the *creative situation*, and the *creative product*. It seems, however, natural that a theory of creativity should be built around an understanding of the creative process, whereas other aspects, although important, have a secondary and deducible nature.

The literature both special and popular abounds in definitions of creativity. As far back as 1959, Taylor surveyed about one hundred definitions in his attempt to clarify the creative process [21]. The definitions vary significantly by the content and the complexity. Nevertheless, there are two commonly recognized 'universal' attributes of creativity: novelty and appropriateness. For the purposes of this study, we will consider creativity as *a cognitive process that generates solutions to a task, which are novel or unconventional and satisfy certain requirements*.

Creativity is not synonymous with intelligence even if in some cases, a correlation can exist between tests of intelligence and tests of creativity [22]. Education and experience may influence creativity, but they are never decisive (either positively or negatively) for the ability to create, too [23]. All the same, there are some inborn, pre-established and individual dispositions, which can be associated with the *thinking style* – not an ability, but a preferred way of expressing or using one or more abilities, that are conducive to the development of creativity [24].

We will distinguish between phenomena of creativity and innovation: the latter normally assumes that a solution to the problem has already been generated, while the former does not require such assumption and is a more general phenomenon. This statement may demolish the popular image of creativity as just 'finding novel combinations of old ideas' or 'having new eyes but not new things' that has been advocated by many authors over the years. In the following, we will argue that creative thinking cannot be reduced to mere search and optimization. Rather, creativity is supported by and makes use of search.

Although being often mystified, it is presently well recognized that creative thinking is a systematic process *successively* running through several phases. According to Wallas' classical model [25], creative thought has to pass through the *preparation* (when the problem is determined), *incubation* (when the unconscious mechanisms work through the problem), *insight* (when ideas of a solution are realized), and *evaluation* (when the ideas are elaborated, tested, and assessed) phases to develop the creative product.

There are two main types of creativity [14]: 1) *improbabilist* that assumes that nothing has to be created *de novo* but existing elements are brought into a distinctive relation to each other by establishing new connections among them, and 2) *impossibilist* – a deeper type that is based on transformation of conceptual spaces. The difference between these types is determined by the mode of creative thinking. Improbabilist creativity stipulates thinking in

-9-

the *associative mode*, adherence to rules, logic, and boundaries of the current *conceptual (mental) space* that is a conceptual packet or network built up for purposes of *local* understanding and action [26]. Impossibilist creativity is subject to the *bisociative mode*, in which the conceptual space is transformed, yet frequently regardless of the existing rules and disciplinary boundaries [27]. As emphasized in [28], a theory of creativity is to be a theory about the exploration, mapping, and transformation of conceptual spaces. It is presumed that a product of impossibilist creativity cannot be generated without *transformation* of the corresponding conceptual space.

By assessing a product of the creative process, two further senses of creativity can be distinguished [14]: *psychological* and *historical*. Psychological (or 'personal') creativity assumes that the product, tangible or intangible, is novel and valuable at least for the agent, who created it. Historical creativity requires that none else has ever produced such a product before. In the latter case, the utility of the product needs to be appraised by a social group the agent belongs to. Some authors, however, maintain a narrower view: in order for a product to be creative, it must be unique to the creator only and must satisfy the creator's criteria of purpose and value [29].

There are two major low-level cognitive mechanisms of creativity [30]: *divergent thinking* that is the ability to generate original, distinct, and elaborate ideas, and *convergent thinking* that is the ability to logically evaluate and find the best solution among a variety of feasible alternatives. A parallel between these two mechanisms and the processes of synthesis and analysis of conceptual spaces can naturally be drawn: creativity *per se* is a balance of synthesis, which usually leads to expanding the conceptual space, and analysis, which is associated with exploring the space [28]. The dynamics of the interaction of divergent and convergent thinking establishes the *canonical dynamics* of the creative process – the dynamics of creative *exploitation* and *exploration*.

Creativity has two levels of dynamics – 'dimensions' – with which one can create: the *system level* and the *domain level*. These dimensions are not strictly orthogonal but rather interrelated so that the former gives a methodology (or structure, style, genre), and the latter determines the application (or presentation) area. The system level is to define a particular medium or particular process, whereas the domain level is to introduce conceptual contents for that, which the medium describes. Creativity is also classified into real-time (or improvisational) creativity, which demands engendering solutions in a short interval of time, and multistage creativity, which lasts as long as necessary to obtain a creative solution.

Domain-level creative *transformations* of the conceptual space most typically include elementary operations of 'negating a postulate' and 'dropping a postulate' [14]. Imposing a (new) context on an already found solution forms another large group of the domain-level transformations.

At the system level, three types of creativity can be distinguished, based on whether *serendipity*, *similarity*, or *meditation* activates the generation of creative ideas. Among these types, 'similarity-based' creativity attracts the most research interest, because *a*) it appears to be understood better and easier, *b*) the study of similarity-based creativity has plain implications for many practical fields of human activity (including the field of engineering design – see [31]), and *c*) there are no reasons to think that the internal processes responsible for generating creative ideas are different for each of the groups (rather, the initial conditions and, as a consequence, the manifestation of the processes differ).

The mechanisms of similarity-based – *analogical* and *metaphorical* – creativity can be referred to as either '*juxtaposition of the dissimilar*' or '*deconceptualization*' [32]. The underlying idea of the former is to put dissimilar concepts or objects so as to find new perspectives and create new meanings through their synthesis, moving away from similarity. The premise is that relaxation of routine associations would force one to reconsider the objects and synthesize new meaningful connections. The theory of cognitive dissonance [33] states that simultaneously considering two or more discording or disharmonizing conceptualizations provokes into a detrimental state of motivation. A typical human reaction to such a cognitive conflict is to try to change attitudes to the objects concerned and/or to obtain more information about the objects, reconstructing or reinterpreting available information and commuting routine associations. Opposing, the 'deconceptualization' mechanism preserves the associations but requires estrangement from the existing conceptualization and describing the object as if it is seen for the first time. New, spontaneous associations can then be originated, and there are two principal ways to create a new structure: '*bottom-up*', when separate concepts are combined into a new whole without any systematic guidance from outside (this may be related to serendipity as well), and '*top-down*', when an external pattern (that can be a metaphor, an analogy, or a methodological template) – *vehicle* is adopted to guide creation. These correspond to hierarchical (bottom-up: by mutating and blending ideas) and heterarchical (top-down: by borrowing and adopting ideas) aspects of handling associative networks of concepts [34].

The understanding of the similarity-based mechanisms clarifies the importance of associations for creative thinking (also see [35, 36, 37]). Generally, the creative process has two potential sources: 1) the consciousness of the creator to activate the necessary abilities and cognitive mechanisms, and 2) the unconscious generation of competing associations – creativity intensively makes use of concepts kept in memory and bound by associative relations [34]. The nature of these relations is neither logical (e.g. causal or inferential) nor mathematical (e.g. transitive or symmetrical), but *perceptual*, *experiential*, and *environmentally induced* (e.g. metaphoric, resemblance, or transformational). (Note that after a creative product has been generated, the associations involved are often reinterpreted in such a way that they acquire a logical or mathematical meaning.) The more 'remote' (in the associative network – conceptual space) elements are used in the process of creation, the higher creative value of the product promises to be. The form (particularly – the 'flatness') of the associative network may be characteristic of the individual's creative potential [35].

There is uncertainty inherent in creative thinking [38]: given a problem represented (understood) in (terms of) a conceptual space, there is usually a family of feasible solutions (or actions), and the final choice among them is determined neither stochastically nor by the input of information from outside, but depends on the intrinsic dynamics of the neural mechanisms. Besides, the conceptual space receives an additional *subjective dynamics* that increases the attractiveness of the chosen alternative and decreases the attractiveness of the rejected alternatives every time after a decision has been made [39]. All this makes unpredictability an essential attribute of the creative process [28]. Although it is often claimed that a well-structured medium allows for faster development and better understanding of the creative product, unpredictability imposes a constraint on the admissible degree of the structuring: as soon as one would be able to know in advance from the structure what the product is going to be in details, the generated product would never be creative *par* *excellence*.

Resuming the above review, we can conclude that creative thinking is a progressive, purposeful, and finite process in the sense that it is sequenced in time from an initial concept to the creative product. On the other hand, it is nonlinear and continual: cognitive processes are hardly ordered, and human thought freely moves from one aspect of the problem to another, regardless any consciously fixed 'cut-off' date. Naturally, the creative process is not always successful in the sense of yielding the product, and its objectives may be abandoned or redefined arbitrarily. Creative thinking is unpredictable in respect to the future, very subjective in terms of the present, and apparent, well interpretable, and even logical if observed retrospectively. It is understood that to computationally deal with the creative process, one first needs *a*) to specify a conceptual space and *b*) to define its dynamics and the basic 'creative' operations – *transformation*, *exploration*, and *exploitation* – on the space. The next and more difficult task would be to identify and manipulate properties or qualities, which make the product creative.

## 3    Modeling the creative process

### 3.1    The domain level: conceptual space, dynamics, and non-determinism

Adopting the notation of Algebraic Semiotics [40], we will consider that a sign system is represented as a five-tuple $\Xi = \langle S, V, C, R, A \rangle$, where $S$ is a sort-set for signs in the system, $V$ is a sort-set for data, $C$ is a set of operations called 'constructors' that are used to create signs from other signs, $R$ is a set of relations defined on the system signs, and $A$ is a set of axioms that constrain the possible signs. $S$ and $C$ are partially ordered: by subsort (denoted '≤') and by level, respectively; in turn, constructors are partially ordered by priority within each level.

We will call a *semiotic morphism* $\mu: \Xi \to \Xi'$ a *translation* (*transition*) that consists of partial functions, which map sorts, constructors, predicates and functions of a sign system $\Xi = \langle S, V, C, R, A \rangle$ to sorts, constructors, predicates and functions of a sign system $\Xi' = \langle S', V', C', R', A' \rangle$ while retaining some of the structure of $\Xi$. The mapping of sorts to sorts preserves the subsort ordering and does not change data sorts, and arguments and result sorts of constructors and predicates are also preserved; more formally [40]:

if $s \leq x$, where $s, x \in S$, then $\mu(s) \leq \mu(x)$;

if $c: s_1 \ldots s_n \rightarrow s$, $s_1 \ldots s_n \in S$, is a constructor (or function) of $\Xi$, then (if defined) $\mu(c): \mu(s_1) \ldots \mu(s_n) \rightarrow \mu(s)$ is a constructor (or function) of $\Xi'$;

if $\pi: s_1 \ldots s_n$ is a predicate of $\Xi$, then (if defined) $\mu(\pi): \mu(s_1) \ldots \mu(s_n)$ is a predicate of $\Xi'$; and

$\mu$ is the identity on all the data sorts and operations in $\Xi$.

A semiotic morphism $\mu$ is called:

– *level preserving* iff whenever sort $s$ is of lower level than sort $x$ in $\Xi$, $\mu(s)$ has lower or equal level than $\mu(x)$ in $\Xi'$;

– *priority preserving* iff whenever constructor $c$ has higher priority than constructor $y$ in $\Xi$, $\mu(c)$ has higher priority than $\mu(y)$ in $\Xi'$; and

– *axiom preserving* iff for each axiom $a$ of $\Xi$, $\mu(a)$ is a corollary of the axioms in $\Xi'$.

A key hypothesis of our study is that human cognition can be characterized as a structuring of experience and perception to provide structured information – a (not necessarily verbal) language – that is to deal with an object (e.g. a product of the creative process). A concept (i.e. object/product model) can be seen as a text 'written' in such a language with a *syntax*, which constrains the object's topological organization, *semantics*, which defines the functional structure and object-environment interaction, and *pragmatics*, which manifests physiological, psychological, and sociological effects associated with the object. This language is composed of signs and is, in itself, a *sign system* that influences and, conditional on the usage context, determines the meaning conveyed with signs.

No truthful description of reality can be made outside the limits of human perception. The creative product is perceived through its *distinctions*, which are revealed as technologic, contextual, and ergonomic relations between product parts and between the product and its environment (natural, social, technical, etc.) [41]. These relations are represented in a design space that is naturally defined as a language. Every time, the language (i.e. a sign system used) may be different but has a common ground – the reality – that fundamentally constrains the relations (but not signs!), if human perception is assumed fairly uniform and consistent.

The language does not prescribe a model, but instead provides with a universe of models for physical and mental phenomena of reality. This allows for the existence of different views while considering the same object.

Given a sign system corresponding to a design space, its elementary sorts (which, together with data sorts, are to represent concepts and concept attributes) and basic constructors and axioms are elicited from the domain general theory (e.g. mechanics, semiconductor circuits, cutting, hydraulics, and the like theories) and from 'common sense' knowledge. These elements constitute the core of *ontological assumptions* about the design problem. The rest of the constructors and axioms determine the design problem specific requirements (e.g. functional solutions and economic constraints), while relations and functions defined in the system give a context for the design (e.g. a technology to be used or intended product-environment interactions).

The language – sign system/conceptual space/design space – allows for representation of the product through its distinctions at both ontological (mereological, topological, morphological, etc.– reflected by the *semantics*) and epistemological (psychological, aesthetic, linguistic, etc.– revealed in the *pragmatics*) levels of description, in any stage of product development. Trivially, an initial product concept is nothing but a composition of related signs. The *natural* (i.e. life-cycle) evolution of the product concept can be described in terms of transitions between different relation patterns by means of a *distinction dynamics* [42, 43].

In designing, the distinction dynamics is driven by the interaction of the processes of *variation* and *selection* that seeks to develop an invariant distinction and constrain the variety of the current relation pattern. The variety of a design of a product intended to operate in an environment is determined by the devised (conceived-represented-realized) product structure (i.e. the relations established between product parts) and the possible relations between the product and the environment (i.e. the product feasible states), which together aggregate the product possible configurations. The variety is defined on and in terms of the product language that includes elements for description of both the structure and the environment. Generally, no variety change can be predicted in advance, and to cope with the changing (at the physical and psychological levels) of the distinctions, the distinction dynamics activates the evolution of product language that thus reflects these dynamics. This evolution is controlled by variation that goes through different configurations of the product and eventually discovers (by selection at every stage of the product life cycle) configurations, which are stable. A constraint on the configurations is then imposed, resulting in the selective retention that decreases the variety but specializes the product language so that only configurations fitting to the environment (i.e. stable relations) ultimately remain. Fundamentally, the design (creative) process always leads to decreasing the number of possible relations between the product and its environment.

To formally deal with the dynamics of the creative process, let us define the following equation (called *Basic Semiotic Component*):

$$P_k(\mu_n: f_n [\Xi_n] \to \Xi_{n+1}), \tag{1}$$

where $\Xi_n$ is a sign system corresponding to an understanding (representation) of the problem at time $t_1$, $\Xi_{n+1}$ is a sign system corresponding to an understanding of the problem at time $t_2$, $t_2 > t_1$, $f_n$ is a composition of semiotic morphisms that specifies the interaction of divergent and convergent thinking (i.e. 'synthesis' and 'analysis'), $\mu_n$ is a semiotic morphism that represents a translation from $\Xi_n$ to $\Xi_{n+1}$, and $P_k$ is the probability associated with $\mu_n$; $\Sigma P_k = 1$, $k=1,\ldots,M$, where M is the number of feasible transformations of the resultant sign system after $f_n$.

$$f_n [\Xi_n] = \Delta^-(\Delta^+(\Xi_n)), \tag{2}$$

where $\Delta^+(\Xi) = \Xi'$ is the *semiotic divergence* operator:

if $\Xi = \langle S, V, C, R, A \rangle$, then $\Xi' = \langle S \cup S', V \cup V', C \cup C', R, A \rangle$, $S' \neq \emptyset$ and/or $V' \neq \emptyset$, and $C' \neq \emptyset$;

$\Delta^-(\Xi') = \Xi''$ is the *semiotic convergence* operator:

if $\Xi' = \langle S', V', C', R', A' \rangle$, then $\Xi'' = \langle S', V', C', R' \cup R'', A' \cup A'' \rangle$, $R'' \neq \emptyset$ and/or $A'' \neq \emptyset$.

There is a partial ranking – *importance ordering* – on the constraints of A in every $\Xi_n$, such that lower ranked constraints can be violated in order for higher ranked constraints to be satisfied. The morphisms of $f_n$ preserve the ranking.

The equation (2) specifies the (semiotic) manifestation of the creative *exploitation* and *exploration* cognitive mechanisms, whereas $\mu_n$ in (1) permits us to describe a transition – creative *transformation* – from one stable state of the problem representation to another stable state. A stable state here corresponds to (partial) closure upon the problem description – the situation, when no external elements need to be added to the current sign system to evolve a solution with the *f*-type morphisms.

Now, we can formulate laws of creative semiosis as follows.

*I. For both the system and domain levels of creativity dynamics, the creative process is represented as a sequence of Basic Semiotic Components.*

For the *present* (i.e. an on-going creative process), there exist a probability distribution over the possible $\mu_n$ for every basic component in the sequence (note however, that due to the *subjective dynamics* of the sign systems – see Section 2 –, the probability distributions cannot be treated as independent). For the *past* (i.e. a completed creative act or process), each of the distributions collapses to a single mapping with $P_k=1$, while the sequence of basic components is degenerated to a (sequence of) function(s). For the *future*, the creative process can be considered in a very general probabilistic sense only (e.g. in terms of thinking styles – by defining probability distributions for $\mu_n$, which are characteristic of a specific domain, social group, design paradigm, and so on).

Let ε be the number of relations between the product and its environment represented in Ξ. We will call a morphism $\mu: \Xi \to \Xi'$ *natural* iff ε > ε′, where ε′ corresponds to Ξ′. Let us also define an *observation* (*interpretation*) operator E as an *axiom-preserving* morphism: E: $\Xi \to \Xi''$. (Most generally, E can be viewed as a mapping representing the effect of an interaction between the product defined in Ξ and its environment.)

*II. A design process represented by Π: $\Xi \to \Xi'$ a (retrospectively generalized composition of) semiotic morphism(s), where Ξ is a sign system corresponding to an initial conceptual space in the beginning of the creative process, and Ξ′ is a sign system in which the creative product is described, is creative if it is a) natural, and b) unique in the sense that there exists no other Π′: E(Ξ) → E(Ξ′).*

(From a more traditional for AI perspective – in terms of product model and considering the *past* that is, obviously, a particular case of the first law of creative semiosis, the creative process can be defined as a mapping

$$f_n[\mu_n(f_{n-1}[\mu_{n-1}(f_{n-2}[\ldots\mu_1(f_0[\Xi_0])\ldots])])] \to \Omega = \langle \Phi, \Lambda \rangle$$

representing a progressive transition from a state described in $\Xi_0$ as an initial specification, to $\Omega$ a model of the creative product – a solution that comprises a requirement specification $\Phi$, that is a formal theory, and its realization $\Lambda$, that is a model of the theory. In other words, $\Omega$ is a representation of the design in the design space engendered by and in terms of $\Phi$; see also [44].)

One can see that the above formulated laws reflect all the main properties of the creative process discussed in Section 2 and specify the domain-level canonical dynamics of the process. Besides, many implications important for computational modeling can be drawn from the first law through interpretation of experimental results. For instance, it has been found [34] that creative thinking is an oscillating process (chaotic for ill defined or ill understood problems and periodic for problems with fixed specification), where the oscillation period is inversely proportional to the amount of constraints associated with the problem and the complexity of problem context. By (2), the total product (i.e. the absolute result) of $\Delta^+$ and $\Delta^-$ determines the content and size of $A$ and $R$. Hence, any details of the internal structure of $f_n$ need not be specified and can be omitted from consideration, if the processing time associated with $f_n$ is directly proportional to the amount of constraints and relations involved. It has also been found experimentally [45] that in designing, the rate of generating creative solutions in a fixed interval of time strongly correlates with the observed dynamics of design space transformations (translations). Then, the computing time of (1) $T = T_f + T_\mu \approx T_f$, as $T_\mu \ll T_f$ (here, $T_f$ is the time required for processing the '$f$-type' morphisms, and $T_\mu$ – for processing the '$\mu$-type' morphisms). The latter imposes a restriction on the complexity of $\mu$.

Some of the theoretical implications of the second law of creative semiosis will be discussed in Section 4, while [41] accounts on the law's technological implication.

### 3.2 The system level: analogical and metaphorical reasoning

Elaborating the ideas originally formulated in [46] and further developed in [47], a model of creative reasoning based on analogue search, adoption, and conceptual blending in engineering design was proposed in [48]. The main hypothesis of the last work is that most (if not all) products are not designed from a *tabula rasa*, but an appropriate analogue or metaphor serves as a vehicle of creative insight and inspiration during designing.

By the model, the analogue-based creative process begins from realizing the objectives and embraces four stages: search of potential solutions, analogue matching, conceptual blending, and reinterpretation. During the first stage, two conceptual spaces are formed: the '*target*' domain that is to arrange candidate solutions satisfying the design functional requirements, and the '*source*' domain that is to evoke and accumulate analogues and metaphors, which are structurally compatible and associatively (conceptually) connected with the problem under consideration. To configure elements of the target domain, target concepts are matched with concepts of the source domain, and the source structure is detailed if no counterpart is available in the analogue chosen. The concepts bound by the analogical relations are 'blended', and another conceptual space is synthesized so that the source concepts create a 'mould', and the target concepts – a 'priming' for the blend. The concepts of the newly created space are then re-interpreted and verified in respect to the problem specification.

Let $\Xi_0$ be a representation of a conceptual space corresponding to the problem initial formulation, $\Xi_{target}$ and $\Xi_{source}$ be representations of the target and source domains respectively (that can be sub-spaces of one conceptual space), and let $\Xi_{blend}$ be a representation of the synthesized conceptual space, with which to describe the creative product – a design. In terms of sign systems, analogical (metaphorical) reasoning evokes transformations of the specified conceptual spaces as follows (for the sake of simplicity, we will limit consideration to the case of the *past* by assuming $P_k = 1$ for each of the semiotic components in the sequence):

1. $\mu_1: f_0[\Xi_0] \rightarrow \Xi_{target}$, $\mu_1': f_0[\Xi_0] \rightarrow \Xi_{source}$,
2. $\mu_2: f_1[\Xi_{target}] \rightarrow \Xi_{blend}$, $\mu_2': f_1[\Xi_{source}] \rightarrow \Xi_{blend}$,
3. $f_2[\Xi_{blend}]$.

Here, $\mu_1$ and $\mu_2$ are *axiom preserving*, and $\mu_1'$ and $\mu_2'$ are *level-* and *priority preserving* semiotic morphisms; $S_0 \subseteq S_{source}$ or $S_0 \cap S_{source} = \emptyset$ that is the *condition of the structural compatibility*. (It is important to note that the operations '$\subseteq$' and '$\cap$' are defined on hierarchical, partially ordered constructions, i.e. on sign systems and their elements [49].

Another important issue is that analogical/metaphorical reasoning generates a conceptual space, which has its own dynamics specified with $f_2$, rather than a single model [47].) The described pattern is valid for utilization of *structural* and *orientational* metaphors. However, to specify deploying *ontological* metaphors, a different scheme still needs to be found.

It should be stressed that the approach presented in this section is to explicate the creative process as the mind manifests it in semiosis, but not to model the actual (physical and chemical) processes in the brain. The described theory is the result of an extensive analysis of many examples of creative acts in different areas of human activity as well as knowledge and experience accumulated by different disciplines in studying creativity.

**3.3     Examples**

The creative potential of analogical metaphorical reasoning in engineering is well illustrated by a classical story about the invention of the telephone by Alexander G. Bell. The devised design of this electronic device combines two independent groups of concepts into one new whole so that new functional solutions from the electromagnetic waves theory are imbedded into the original structure of human ear. The electric signal plays the rôle of and substitutes for the medium for transmitting information, and knowledge of the functioning of the inner ear membrane prompts the application of a moving conductor in the speaker and microphone. While such an 'obvious' analogue might drive the creative process 'from scratch', it is a historical fact that Bell's idea of the telephone design was inspired by another invention misinterpreted by Bell as if the reported device had the function of a telephone: J.P. Reis constructed a rudimentary transmitter by placing an animal ear membrane in front of an electrical contact; a galvanic inductor oscillated in the receiver in the same manner as the transmitted signal. Whether only the analogue with ear guided the creation or the design emerged through the evolution of a conceptual space by dropping/negating constraints and imposing a different context, the adopted structure became the key to the creative solution after the problem had been understood, and the corresponding initial conceptual space had been constructed. One would naturally interpret the creative act in terms of sign systems, which represent the conceptual spaces, and semiotic morphisms (see Figure 2).

There is plenty of room for creative analogical (metaphorical) reasoning in everyday engineering work. Engineering findings may not be revolutionary for the science and society, but they are necessarily useful and usually well documented. Figure 3 presents a simple example of the application of the semiotic model of analogical reasoning in the development of a medical computer system.

A completed design of a hardware/software system for telemetering the heart rate has been used as an analogue during designing another system intended to analyze the direct-current resistance of skin at different places of the body. The original system (the *source*) includes a differentiating amplifier to obtain the electrocardiogram-type signal (ECG), a filter that filters out the segment R of ECG (R can be used as the mark of a heartbeat), an interface module, and a PC. The system software reads the signal from the serial port and provides recording 'logic0/logic1' a time series that reflects the change of the heart rate over time. The *target* domain was initially composed of such concepts as PC, meter, and interface. Figure 4 illustrates how the association is built through the subsumption relation, and how the analogue is sought. Figure 3 shows the development of the new system design by blending the source and the target concepts. The solution found is simple, inexpensive, reliable, and has a low metering error and a high modularity.

More historical as well as of the authors' design practice examples of the specification of system-level creativity dynamics can be found in [50] and [51].

## 4  Design + emergence = creative value?

The unpredictability intrinsic to a creative act is often associated with emergence [52]. In the recent years, there have been extensive efforts made in studying emergence that, however, have not yet yielded a comprehensive theory, a reasonably universal explanation or, at least, a precise definition of the phenomenon (see [53]). An overwhelming majority of the definitions proposed in the literature just highlighted certain aspects of emergence most appreciable and/or interesting in a given domain. In design, emergence may be related to the generation of a structure, configuration, or pattern that constitutes a functional unit with properties not derivable by mere summation of its parts. Because of the not well understood but evident relationship between creativity and the phenomenon of emergence, it appears interesting to speculate as to if and why emergent properties have an effect on the creative value of the product.

In his attempt to formulate the necessary condition of emergence, Heylighen [54, 42] identified four principal processes in the development of a complex system: *internal variation*, *external variation*, *internal selection*, and *external selection*. Variation stands for the generation of a variety of simultaneously present, distinct systems or of subsequent, distinct states of the same system by way of recombination of already existing distinct entities (systems or system attributes). Internal variation makes use of the entities belonged to the system under consideration, while external variation acquires the entities for recombination from outside the system. Selection can be thought of as the elimination of certain distinct entities (systems or states). For internal selection, a selection criterion depends on the internal structure of the entity, and external selection is subject to a criterion defined on the entity environment. Heylighen defines emergence as 'the transition after variation from a given system to a different, selectively retained system, characterised by a new (partially) closed model' [54]. His *necessary condition* of emergence contends that if variation and selection processes operate independently, the result of their interaction cannot be predicted and should be determined emergent. To clarify the phenomenon of emergence in creative design, we will interpret the above definition and the necessary condition in terms of the semiotic model of the creative process. Our interpretation fits into the philosophical framework proposed in [55], and it provides for epistemological (by taking into account the rôle of the observer) and ontological (by including the metaphysical assumptions prior to the act of observation) perspectives in dealing with emergence.

Interpreting the second law of creative semiosis (see Section 3.1) in the light of the semiotic definition of emergence from [20], we can postulate that

*a property p is emergent iff $p \in E(\Xi')$ and $p \notin E(\Xi \cap \Xi')$, where $E(\Xi') \neq \emptyset$.*

The emergence is deducible or computational if there can be determined (e.g. through logical inference, computation, or exhaustive search) a morphism $E(\Xi) \rightarrow E(\Xi')$; otherwise the emergence is observational.

By applying Heylighen's general definition, the design creative process reveals three potential sources of emergence: the interpretation E, the (sequence of) morphism(s) Π, and the ontological assumptions underlying the construction of a particular Ξ.

In the first and, perhaps, most usual case, emergence appears a purely psychological and observationally relativistic issue that imposes no epistemological requirements on the corresponding semiosis processes. An increase of knowledge about the observable phenomena can disclose that the property 'anomalies' of the 'whole' can be explained in terms of the 'parts'. Handling emergence is then a matter of instantiating correct interpretations that is beyond the creative process. In this case, emergence may be related with creativity only at the personal, highly subjective level. The necessary condition of emergence is easily satisfied if the processes of design (seen, for instance, as variation) and interpretation (seen as selection) are independent.

If the property $p$ cannot be described in terms of $\Xi \cap \Xi'$, but there exists (or could principally be built) a different $\Xi''$ such that $p \in E(\Xi \cap \Xi'')$, emergence is a *(by-)product* of the creative process, and it essentially depends on (the properties of) the particular morphism $\Pi: \Xi \to \Xi'$. To detect the emergent property, there must be allocated at least two levels of description accountable for the creative product, e.g. the levels of 'parts' and 'wholes' (or the 'target' and 'source' and the 'blend' levels in the model of analogical reasoning – see the previous section). The ontological assumptions of $\Xi$ are supposed to be fixed during creation, and the relations between the levels are not changed over time. Hence, the levels can be treated independently that ensures the possibility of emergence. Creativity by finding novel combinations (selection at the level of 'wholes') of old ideas (variation at the level of 'parts') is powered by this type of emergence. However, the singularity of the creative product naturally dwindles as $\Xi''$ is evolved, reducing the significance of the specific Π and disclosing the effects of the upper level in terms of the lower levels.

When the property $p$ cannot be described with $\Xi$ inasmuch as it contradicts with the sign system's very organization and structure (e.g. $p$ does not conform to the priority ordering of the constructors of $\Xi$ which, for instance, represent the causal structure of the reality), the source of emergence is the ontological assumptions represented in $\Xi$. In reality, $\Xi$ is hardly fixed during creation, and the design creative process can result in emergence due to the specificity (fallacy, incompleteness, inconsistency, etc.) of the embraced ontological assumptions that may later be revealed as an ontological category error. Again, there are at least two levels necessary to realize the emergent property, but in this case, they are not independent – changing $\Xi$ can result in changing the relations between the levels, just as detecting stable emergent patterns at the upper level can lead to imposing new constraints and changing the organization of the lower levels that, in turn, can affect the upper level. The singularity (subjectivism) of the ontological assumptions underlying the design (at the level of parts) provides for variation that is, in general, independent of any natural process of selection (at the level of wholes) defined on the product environment.

Thus, at this point of the discussion, we can conclude that only in the second case (i.e. when the source of emergence is the creative process itself), the emergent property can strongly be associated with the creative value of a product in engineering design. As long as a particular design process is still unique, the designed product has good chances to be creative if it has some utility at all. Interpretation as the source of emergence has nothing to do with the 'objective' (not psychological) creativity, and the singularity of the ontological assumptions underlying a design may reduce the product's utility (e.g. because of problems in product-environment interaction) rather than increase its creative value. (In the latter case, however, the emergent property very often determines the creative value in art.)

## 5    Concluding remarks

History bears witness that creativity is crucial for the progress of civilization. In the reported study, we explored human creativity in engineering design, while particularly focusing on the semiotic aspects of the creative process. We also investigated the mechanism of analogue-driven creative thought and, in an attempt to understand what makes the product creative, showed how the phenomenon of emergence can be interpreted from a semiotic perspective. It was the purpose of this study to develop a realizable approach to modeling creativity-related phenomena in the design process, to promote a better understanding of creativity, its mechanisms, dynamics, and manifestation. The main argument of the research is that computational modeling of the thought processes does not necessarily require modeling physical processes. Rather, it could be done through modeling the corresponding semiosis processes.

This paper offers one new contribution: for the first time in the domain, the dynamics of the creative process has unambiguously and, in fact, computationally been described. Besides, the presented research findings are based upon and strongly supported by the literature, so that this paper contributes to systematization of design knowledge and provides the reader with a thorough introduction to the modern study of creativity. The former would educationally be useful for the planning of design courses, and the latter would, hopefully, encourage the engineering community to a broad discussion of creativity phenomena.

The presented theoretical materials lay the initial basis for a new computational theory of design that is the main result of this study. It must be noted, however, that although the principles of creative design have been formulated with sufficient (for computational purposes) precision, it is recommended to apply them (alike Algebraic Semiotics in general) in an informal way, calling for details only in boundary and difficult situations. The main purpose of these as well as other not-yet-formulated semiotic laws is not to invent 'creative computations' but to guide the examination of product development and usage processes, no matter which design theory or even paradigm is employed.

The work has a few implications for practice. By the proposed model of creative analogical (metaphorical) reasoning, inspiration and creativity in design acquire quite a realizable and pragmatic meaning: inspiration is reduced to knowing *what* to adopt and borrow from previous experience, and creativity becomes in much a matter of *how* to reuse that which is borrowed. This implies that computer-aided design methodologies could benefit from incorporating relevant services for function- and structure-based information retrieval. Tools and methodologies for the generation and detection of designed product emergent properties would, perhaps, aid the designer in assessing the (potential) creative value of the product.

The authors have been developing a software system that could simulate creative transformations of conceptual spaces to practically explore the dynamics of the synthesis-analysis interaction during designing [50]. It is planned to expand the theoretical basis of the semiotic theory of design so as to clarify the origin and nature of the embedded externality (see [38]) in creative thinking. The latter, however, is difficult to do without understanding of the post-creation dynamics of the conceptual space [56]. Investigation of sufficient conditions of emergence represents another topic for future research that develops from this paper.

**Acknowledgement**

This work has been made within the 'Methodology of Emergent Synthesis' research project (No 96P00702) funded by the Japan Society for the Promotion of Science.

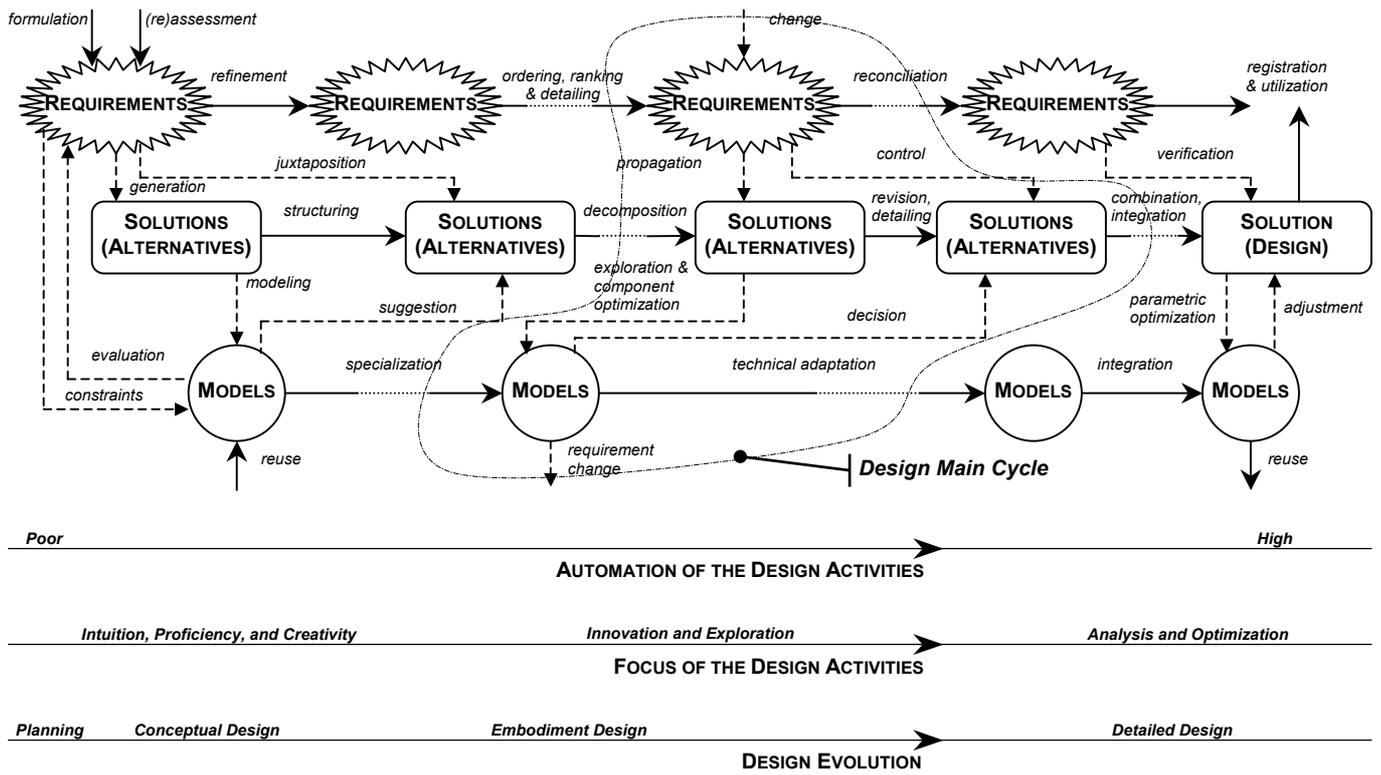

Figure 1.

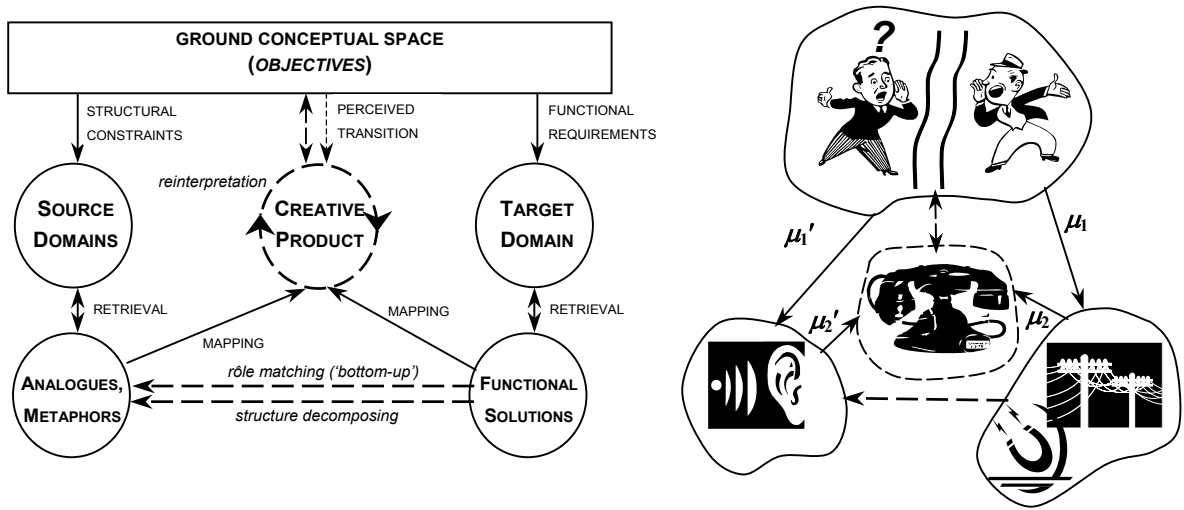

Figure 2.

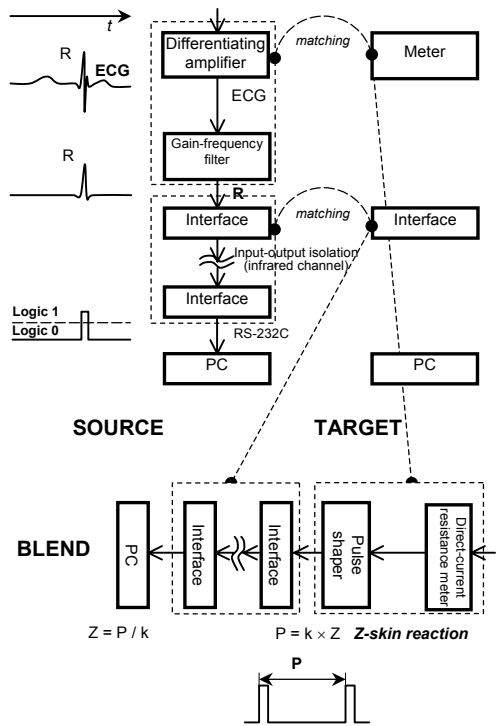

Figure 3.

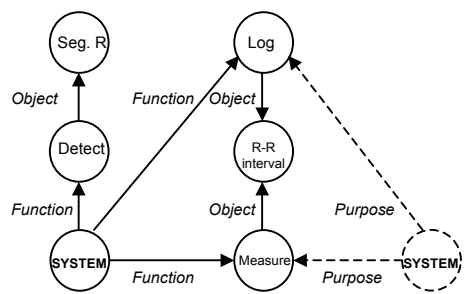

Figure 4.

Figure 1. Computer-aided design.

Figure 2. Creative analogical (metaphorical) reasoning and the invention of the telephone.

Figure 3. Creative analogical reasoning: a practical example.

Figure 4. Building the association.